\documentclass[conference]{IEEEtran}

\ifCLASSINFOpdf
   \usepackage[pdftex]{graphicx}
   \graphicspath{{../pdf/}{../jpeg/}}
   \DeclareGraphicsExtensions{.pdf,.jpeg,.png}
\else
\fi

\PassOptionsToPackage{hyphens}{url}\usepackage{hyperref}
\usepackage{xcolor}
\usepackage{hhline}
\usepackage{multirow}
\usepackage{amsmath,amssymb,amsfonts}
\usepackage{algorithm}
\usepackage{algpseudocode}
\usepackage{verbatim}
\usepackage{balance}

\newcommand{\ignore}[1]{}

\begin{document}
\title{Object Removal Attacks on LiDAR-based 3D Object Detectors}

%on LiDAR 3D Point Clouds\\ for Autonomous Vehicle Perception}

\author{
    \IEEEauthorblockN{Zhongyuan Hau\textsuperscript{\textsection},
    Kenneth T. Co\textsuperscript{\textsection},
    Soteris Demetriou, 
    Emil C. Lupu
    }
    \IEEEauthorblockA{Imperial College London}
    \IEEEauthorblockA{\{zy.hau17, k.co, s.demetriou, e.c.lupu\}@imperial.ac.uk}
}


%Telephone: (800) 555--1212, Fax: (888) 555--1212}


\IEEEoverridecommandlockouts
\makeatletter\def\@IEEEpubidpullup{6.5\baselineskip}\makeatother
\IEEEpubid{\parbox{\columnwidth}{
    Workshop on Automotive and Autonomous Vehicle Security (AutoSec) 2021 \\
    21 February 2021 \\
    ISBN 1-891562-68-1 \\
    https://dx.doi.org/10.14722/autosec.2021.23016 \\
    www.ndss-symposium.org
}
\hspace{\columnsep}\makebox[\columnwidth]{}}

\maketitle

\begingroup\renewcommand\thefootnote{\textsection}
\footnotetext{Equal contribution}
\endgroup

\begin{abstract}
%\boldmath
LiDARs play a critical role in Autonomous Vehicles' (AVs) perception and their safe operations. Recent works have demonstrated that it is possible to spoof LiDAR return signals to elicit fake objects. In this work we demonstrate how the same physical capabilities can be used to mount a new, even more dangerous class of attacks, namely \textit{Object Removal Attacks} (ORAs). ORAs aim to force 3D object detectors to fail. We leverage the default setting of LiDARs that record a single return signal per direction to perturb point clouds in the region of interest (RoI) of 3D objects. By injecting illegitimate points behind the target object, we effectively shift points away from the target objects' RoIs. Our initial results using a simple random point selection strategy show that the attack is effective in degrading the performance of commonly used 3D object detection models.

%We further show that for small objects such as pedestrians and cyclists, the attack is effective at distance of 11m and beyond.
\end{abstract}


% \fi
%%\IEEEpeerreviewmaketitle

\section{Introduction}
We are currently undergoing a revolution in transportation and mobility. New generations of vehicles are increasingly equipped with high-precision depth sensors to better perceive their environment and offer unprecedented levels of driver assistance and driving autonomy. Such vehicles commonly rely on LiDAR sensors, which collect high definition depth measurements stored in 3D point clouds. Reliably detecting objects from such point clouds is vital to the safety of the autonomous vehicle, its users and passengers.

%LiDARs (derived from light detection and ranging) are high-precision depth sensors which are pervasively deployed~\cite{bbc_news_2019, coldewey_2018} in autonomous vehicles (referred to as AVs henceforth) where a new class of Deep Neural Network (DNN) 3D object detectors leverage depth sensor measurements (processed in batches called 3D point clouds) to detect objects -- a necessary task for downstream safety-critical driving decision-making \cite{qi2017pointnet++, PRCNN_Shi_2019_CVPR, Shi2020PointGNNGN, hau2020ghostbuster}.

\vspace{3pt}\noindent\textbf{LiDAR spoofing attacks.} Recent studies have shown that it is possible to attack LiDAR-based perception systems by spoofing LiDAR return signals~\cite{petit2015remote, cao2019adversarial, shin2017illusion, hau2021shadowcatcher}. Petit \textit{et al.} first demonstrated this with physical attacks that can inject up to 10 fake 3D points in a point cloud. Cao \textit{et al.} and Sun \textit{et al.} progressively improved on the physical capabilities of the LiDAR spoofing adversary showing that one could reliably inject up to 60 and 200 fake points respectively. More importantly, Cao \textit{et al.} developed a white-box model-level digitally simulated LiDAR spoofing attack that can introduce front-near fake measurements in a scene, which are then detected as objects by an end-to-end autonomous vehicle (AV) system. Sun \textit{et al.} then demonstrated both white-box and black-box attacks that spoof vehicles in front-near locations by exploiting patterns of occluded and distant vehicles. Xiang \textit{et al.} \cite{xiang2019generating} examined the vulnerability of point-cloud based object detectors and proposed white-box approaches for point shifting and point injection to craft adversarial point-clouds. Zhao \textit{et al.} \cite{zhao2020nudge} proposed a class of point cloud perturbation attacks that minimizes the number of points perturbed to flip the results of point-cloud based object detectors. They use gradient-based and genetic algorithm approaches to generate adversarial point clouds with perturbations of up to 150 points to subvert object detection with a 95\% success rate.

%\vspace{3pt}\noindent\textbf{Adversarial point clouds.}
%\vspace{3pt}\noindent\textbf{LiDAR spoofing attacks.} Unfortunately, recent studies have shown that it is possible to attack LiDAR-based perception systems by spoofing LiDAR return signals~\cite{cao2019adversarial, petit2015remote, shin2017illusion}. Cao \textit{et al.}~\cite{cao2019adversarial} further conducted a study of LiDAR spoofing attacks on AVs and formulated a white-box model-level spoofing attack based on perturbations that successfully spoofed objects and affected downstream AV decision-making. Sun \textit{et. al.}~\cite{255240} demonstrate both white-box and black-box attacks that spoof vehicles in front-near locations by emulating patterns of occluded and distant vehicles. Xiang \textit{et. al.} \cite{xiang2019generating} examined the vulnerability of point-cloud based object detectors and proposed white-box approaches for point shifting and point injection to craft adversarial point-clouds. Zhao \textit{et. al.} \cite{zhao2020nudge} proposed a class of point cloud perturbation attacks that minimizes the number of points perturbed to flip the results of point-cloud based Object Detectors. They use gradient-based and genetic algorithm approaches to generate adversarial point clouds with perturbations of up to 150 points to subvert object detection at a 95\% success rate.

\vspace{3pt}\noindent\textbf{Object hiding.} Tu \textit{et al.} \cite{tu2020physically} proposed both white-box and black-box methods to generate adversarial objects that when placed above a target vehicle, would evade point-cloud based object detectors with a success rate of 80\%. For the white-box attack, the adversarial object is generated using a gradient-based approach to minimize the confidence score of the target object (vehicle). A black-box attack was also demonstrated, where the adversarial objects are chosen using a genetic algorithm approach to iterate and improve adversarial object meshes. Object-hiding attacks are considered more dangerous than spoofing objects. Whilst detecting a spoofed object can bring the ego-vehicle to a full stop, failing to detect an object has a higher chance of leading to a fatal collision.

%All these highlight an impending need for better understanding the capabilities of LiDAR spoofing adversaries and the effect they can have in the safety critical task of autonomous driving. 

%Front-near objects, those that are directly in front of the vehicle at close proximity, are of great concern as any detection failure could result in potentially life-threatening accidents \cite{wakabayashi_2018}. Reliably detecting objects are paramount for the safe operation of AVs and understanding their vulnerabilities can help us design better defenses.

%% TODO: Add IRL motivation of object hiding attack (front-near case, and more general high-speed case)
%\vspace{3pt}\noindent\textbf{Front-near object hiding.}

\vspace{3pt}\noindent\textbf{Our work}. We leverage the demonstrated state of the art capabilities of the physical LiDAR spoofing adversary~\cite{petit2015remote, shin2017illusion, cao2019adversarial} to design a new model-level \textit{object removal attack} (ORA) that aims to hide objects from 3D object detectors. Compared to prior work with spoofed objects~\cite{petit2015remote, shin2017illusion, cao2019adversarial} ORAs have a different goal: the adversary aims not to introduce a ghost object but force mis-detection of a genuine object which can have more severe consequences. Moreover, in contrast with related work on 3D object hiding~\cite{tu2020physically}, we introduce a new technique that does  not aim to introduce patterns on top of genuine objects, but rather to spoof points within a genuine object's bounding box such that they appear away from their original position and cause object mis-detection. ORAs are stealthier since they do not require placing adversarial objects on the target, are easier to mount and have high success rates. We conduct digital ORA attacks, emulating the physics of LiDAR operation. Their effectiveness is evaluated against popular 3D object detectors (PointRCNN~\cite{PRCNN_Shi_2019_CVPR} and Point-GNN~\cite{Shi2020PointGNNGN}).

We found that an adversary with the ability to inject $\leq 200$ points can reduce their \textit{recall} to less than 25\% for \textit{Pedestrian} and \textit{Cyclist} object classes on both models with a \emph{random} point selection strategy. Our work demonstrates the feasibility of ORAs and we hope to inspire future work on more sophisticated ORA strategies that can lead to a better understanding of the LiDAR spoofing adversary model.

%in the general case. Additionally, it can drop the recall for both models to less than 25\% for cyclist and pedestrian objects at a short range of 11-14m.

% \ignore{
% \vspace{3pt}\noindent\textbf{Contributions}. Below we summarize our main contributions:

% \vspace{5pt}\noindent $\bullet$ We propose a LiDAR perception attack that explores using LiDAR point injection to perturb the perceived environment by \emph{shifting points} from a legitimate object's point cloud to prevent it from being detected by a LiDAR-based object detector.

% \vspace{5pt}\noindent $\bullet$ Digital attack that emulates the physics of LiDAR operation to simulate the attack scenarios.
% }
%% TODO: No need paper outline (below) if very short?

% \vspace{5pt}\noindent\textbf{Paper Organization.} In Section~\ref{sec:attack} we discuss background information, the threat model we use and present our Object Removal Attack. In Section~\ref{sec:experiments} perform an in-depth evaluation of the proposed attack's effectiveness and discuss other aspects of our work. Then, in Section~\ref{sec:related_work} we identify and discuss relevant prior work and present our conclusions in Section~\ref{sec:conclusion}.

\begin{table*}[t]
\footnotesize
\caption{Average Precision (AP) of 3D object detection for different classes under the ORA random attack.}
\label{table:combined}
\centering
\begin{tabular}{c|c|ccc|ccc|ccc}
    \hline
	\multirow{2}{*}{Model} & \multirow{2}{*}{Attack Budget} & \multicolumn{3}{c}{Car AP (IoU = 0.7)} & \multicolumn{3}{c}{Pedestrian AP (IoU = 0.5)} & \multicolumn{3}{c}{Cyclist AP (IoU = 0.5)}\\
	 &  & Easy & Moderate & Hard & Easy & Moderate & Hard & Easy & Moderate & Hard\\
	\hline\hline
    & 0 (Clean) & 88.86 & 78.61 & 77.75     & 62.87 & 54.88 & 48.95    & 73.08 & 56.33 & 52.36\\   
    & 10 & 79.30 & 65.09 & 58.11            & 50.69 & 42.96 & 38.15    & 61.03 & 39.04 & 35.58\\
	& 20 & 78.85 & 59.57 & 51.00            & 48.06 & 40.38 & 35.42    & 54.66 & 31.67 & 30.12\\
	\multirow{2}{*}{PointRCNN \cite{PRCNN_Shi_2019_CVPR}}
	    & 40& 77.02 & 54.36 & 46.13         & 43.66 & 35.84 & 31.96    & 41.81 & 24.36 & 22.96\\
	& 60 & 72.97 & 48.01 & 40.42            & 38.70 & 31.97 & 27.82    & 33.54 & 19.14 & 18.64\\
	& 100 & 64.69 & 40.97 & 33.47           & 35.99 & 28.46 & 23.20    & 24.48 & 15.75 & 15.77\\
	& 150 & 55.74 & 34.05 & 28.33           & 29.04 & 22.79 & 19.93    & 17.01 & 11.95 & 11.33\\
	& 200 & 47.32 & 31.05 & 25.00           & 25.62 & 19.02 & 16.92    & 13.07 & 9.43 & 9.42\\
	\hline
	& 0 (Clean) & 89.89 & 88.82 & 87.75     & 73.74 & 70.34 & 63.57   & 85.69 & 64.44 & 62.25\\
    & 10 & 89.62 & 78.33 & 69.31            & 73.74 & 70.25 & 63.49   & 85.69 & 64.17 & 62.01\\
	& 20 & 89.34 & 71.38 & 62.06            & 71.89 & 64.01 & 61.19   & 76.79 & 52.78 & 48.32\\
	\multirow{2}{*}{Point-GNN \cite{Shi2020PointGNNGN}}
	& 40& 86.40 & 62.50 & 53.08             & 63.18 & 55.57 & 48.28   & 57.00 & 36.70 & 34.85\\
	& 60 & 80.26 & 53.71 & 44.60            & 56.27 & 48.10 & 44.42   & 43.45 & 26.98 & 26.04\\
	& 100 & 70.47 & 43.89 & 35.36           & 47.15 & 39.24 & 32.50   & 25.63 & 16.06 & 15.86\\
	& 150 & 57.15 & 34.61 & 29.17           & 39.23 & 31.65 & 27.72   & 10.86 & 8.14 & 7.40\\
	& 200 & 45.58 & 28.01 & 22.01           & 31.99 & 26.64 & 23.81   & 4.76 & 3.66 & 3.85\\
	\hline
\end{tabular}
\end{table*}
\vspace{-2mm}

\section{Object Removal Attack}\label{sec:attack}
%We perform digital attacks on the point clouds obtained in real AV scenarios. We simulate the injection of points behind an object while removing the original point in the same ray direction, mimicking as closely as possible the physics of LiDAR operation in a real world environment.% from the KITTI Dataset~\cite{Geiger2013IJRR}.
We introduce a new class of attacks, namely object removal attacks (ORAs). ORAs are black-box, model-level attacks that can be launched by a physical LiDAR spoofing adversary. The goal of ORAs is to displace the original depth measurements within a genuine object's bounding box to appear outside that bounding box such that the target object is not detected by 3D object detectors. Below we elaborate on our threat model and a preliminary strategy to demonstrate the feasibility of ORAs.

%\subsection{LiDAR Operation for Autonomous Vehicles}
%LiDARs sense the environment by emitting pulses in the invisible near-infrared wavelength (900–1100 nm), which are reflected on incident objects before returning to the receiver of the emitter device. Based on the time of flight, the distance between the sensor and the incident object is calculated. LiDARs used in AVs (e.g. Velodyne LiDARs) emit a number of light pulses from an array of vertically arranged lasers (16, 32, 64, etc.) that rotate around a center axis to obtain a 360-view of the surroundings of the sensor unit. The return signal is translated to 3D point measurements consisting of its coordinates and a reflection value corresponding to the return signal's reflectivity or signal strength. The default setting for measuring return signals for LiDARs is the \emph{Strongest Return Mode}, where only the strongest return signal is registered when there are multiple return signals from the same direction.

\vspace{-1.5mm}
\subsection{Threat Model}
\noindent\textbf{Physical Capabilities.} We assume an adversary $\mathcal{A}$, who can spoof the LiDAR return signals of a target AV~\cite{cao2019adversarial, petit2015remote, shin2017illusion, 255240}. The adversary can achieve this by deploying a device within the line of sight of a victim vehicle's LiDAR sensor. The adversarial device can capture LiDAR signals, alter them and emit them toward the victim sensor with a controlled delay. By controlling the return signal, $\mathcal{A}$ can manipulate the resulting 3D measurements reported in a 3D point cloud by the victim sensor. We assume $\mathcal{A}$ has state of the art sensor spoofing capabilities and can inject $\leq200$ points in a 3D scene within a horizontal angle of 10$^{\circ}$~\cite{255240}. We also assume that $\mathcal{A}$ can spoof resulting measurements to appear further away from the target vehicle than they actually are~\cite{shin2017illusion}.

\noindent\textbf{Digital Capabilities.} Since $\mathcal{A}$ is in physical proximity, we assume it can also sense the environment and detect objects within the vicinity of the target vehicle. Using basic transformations, $\mathcal{A}$ can change the coordinate system of a 3D scene from the reference point of $\mathcal{A}$ to that of the target vehicle.

\vspace{-1.5mm}
\subsection{Object Removal Attack (ORA) Rationale}
%Object removal works by ... (explain mechanism for removal of points) include settings of Velodyne or whatever hardware simulated
LiDAR spoofing has been previously demonstrated~\cite{cao2019adversarial, shin2017illusion, 255240}. These attacks exploit the fact that LiDARs deployed in AVs operate under the Strongest Return Mode setting, where only a single measurement can be recorded per ray direction. Therefore, the injection of a spoofed point results in the original corresponding point in the direction of the ray (from the LiDAR to the spoofed point) to be displaced. We also leverage this phenomenon, but to achieve a different goal. Instead of aiming to introducing objects, we highlight a new class of attacks, namely \emph{Object Removal Attacks (ORAs)} that aim to remove objects. Our proof of concept hides an object from AV perception by displacing points from a target object's point cloud with LiDAR point injection behind that object.

%\subsection{Threat Model}
%We adopt the threat model from~\cite{cao2019adversarial, 255240} and assume a physical adversary who can spoof LiDAR return signals to fool object detection of an AV's 3D object detector model. The adversary can spoof the signals either by placing an attacker device on the roadside or by mounting it on an attack vehicle driving in front of the target vehicle in an adjacent lane~\cite{cao2019adversarial}. The attacker's device can capture the LiDAR signal, and emit a return signal with a delay which controls where in the resulting point cloud the spoofed point will appear. This has been proven as a realistic attack surface by previous work~\cite{cao2019adversarial, petit2015remote, shin2017illusion, 255240}. We next define the adversary's ($\mathcal{A}$) capabilities and goals:

%\vspace{3pt}\noindent$\bullet$ \textit{Number of spoofed points.} We assume $\mathcal{A}$ has the state of the art sensor spoofing capabilities and can inject $\leq200$ points in a 3D scene within a horizontal angle of 10$^{\circ}$~\cite{255240}.

%\vspace{3pt}\noindent$\bullet$ \textit{Aims.} The aim of $\mathcal{A}$ is to inject points (within its budget and capabilities) to cause a target object to evade detection.

\begin{algorithm}[t]
  \caption{Obtaining Attack Trace (ORA-Random)}\label{algo:1}
  \begin{algorithmic}
     \State \textbf{Input:} list($obj\_pts\_coords$) \Comment{Target Object's Point Cloud}
     \State $candidate\_pts\_coords$ =[]
     \State $attack\_trace\_pts\_coords$ = [] 
     \For{\textbf{each} \texttt{pt} \textbf{in} $obj\_pts\_coords$}
        \If{ \texttt{pt} \textbf{within} \texttt{spoofing\_horizontal\_angle}}
            \State $candidate_\_pts\_coords$ $\gets$ \texttt{pt}
        \EndIf
    \EndFor
    \State $attack\_pts \gets$ random($candidate\_pts\_coords$, $\mathcal{A}_{budget}$)
    \State $attack\_trace\_pts\_coords \gets$  ($pts$ in $obj\_pts\_coords$ $\wedge$ not in  $attack\_pts$)
    \For{\textbf{each} \texttt{pt} \textbf{in} $attack\_pt$}
        \State $attack\_pt\_coords \gets$ dist\_increment\_along\_ray(\texttt{pt})
        \State $attack\_trace\_pts\_coords$.append($attack\_pt\_coords$) 
    \EndFor
    \State \Return $attack\_trace\_pts\_coords$
  \end{algorithmic}
\end{algorithm}
\vspace{-2mm}

\begin{figure*}[htb]
\centering
\includegraphics[width=0.85\textwidth, trim={0 2.2em 0 0}, clip]{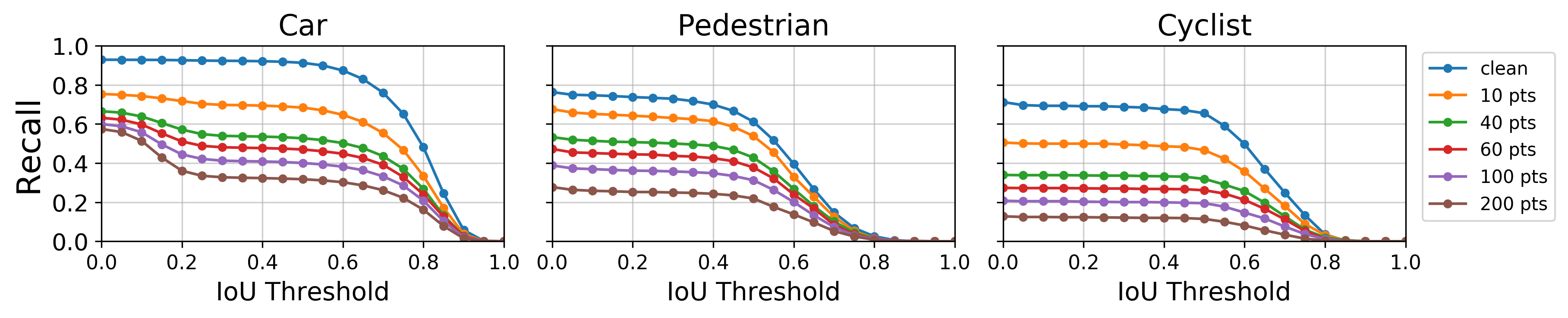}
\includegraphics[width=0.85\textwidth, trim={0 0 0 0}, clip]{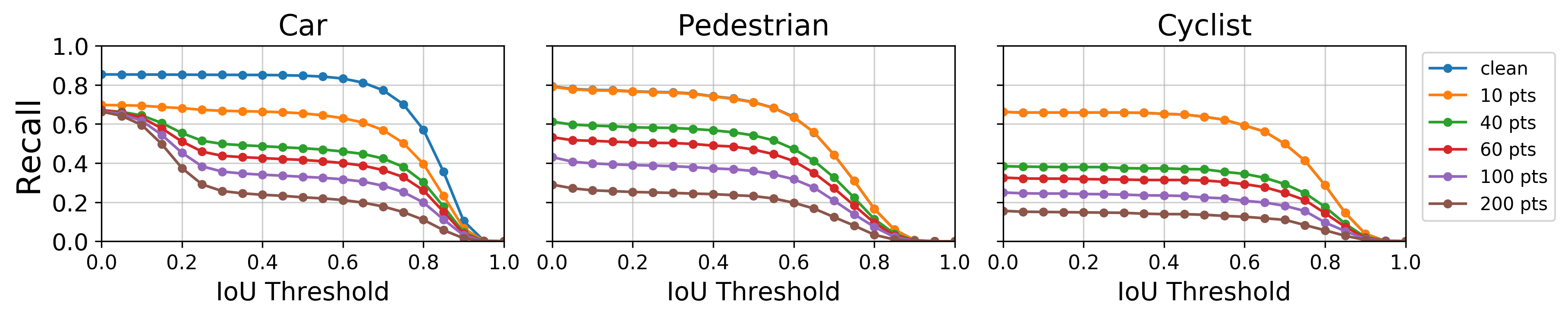}
\vspace{-3.5mm}
\caption{Recall-IoU curves for (top) PointRCNN and (bottom) Point-GNN under the ORA random attack with different IoU thresholds.}
\label{fig:recall_iou}
\end{figure*}
\vspace{-1mm}

\subsection{ORA Operating Mechanism}

% \ignore{
% How do we select points to remove for ORA?
% a) test random
% b) test black-box BayesOpt (justify -- preprocessing of point clouds can often be non-differentiable, difficult to perform gradient descent on vs. traditional attacks)
% Cite -- mesh paper, other papers
% In this work, Object Removal Attack (ORA) is... (just random?)
% and then BayesOpt version, but that's for later??
% Random attack evaluation (describe random attack in detail -- done per object, etc.)
% }

An ORA exploits the LiDAR's default mode for recording the measurements of return signals--where a single return signal per ray direction is recorded. This enables an adversary to perform point injections that can remove points from a target object's original point-cloud by spoofing another signal at a different location in the same direction of the rays that are incident on an object. The resulting perturbed point-cloud would cause point-cloud based object detectors to miss detecting the target object, and thus evading object detection.

%% Add a comment on how the attacker can transform their view to the ego-vehicle's view (transforming attacker's origin to ego-vehicle's origin)

The process of ORA first starts with the adversary identifying the target object's location and the region of interest (RoI) where points are to be selected from and removed. Here, we assume that the adversary has knowledge of the 3D scene and is able to obtain bounding boxes of target objects and the coordinates of the points in the bounding boxes (i.e. object points). This can be achieved by finding a translation matrix that changes the coordinates system from the reference point of the attacker to the ego-vehicle~\cite{ganapathy1984decomposition}.  Next the adversary obtains a set of points from the object points that are within its spoofing horizontal angle (i.e. candidate points); one way of achieving this is to segment the object bounding box by the spoofing angle and only use the points within the segment. In our case, we used the left-most coordinates of the bounding box as an anchor point to calculate the points that are within the horizontal spoofing angle. Lastly from the subset of candidate points within the spoofing angle, the adversary picks points within its budget to perform point injection, spoofing points at a random distance behind the original points' location (in the direction of the ray). ORA is modular and can be used with various point selection strategies. We demonstrate ORA with a random point selection (\textit{ORA-Random}) from the candidate points. \textit{ORA-Random} is detailed in Algo.~\ref{algo:1}.

\begin{figure*}[htb]
\centering
\includegraphics[width=0.85\textwidth, trim={0 2.2em 0 0}, clip]{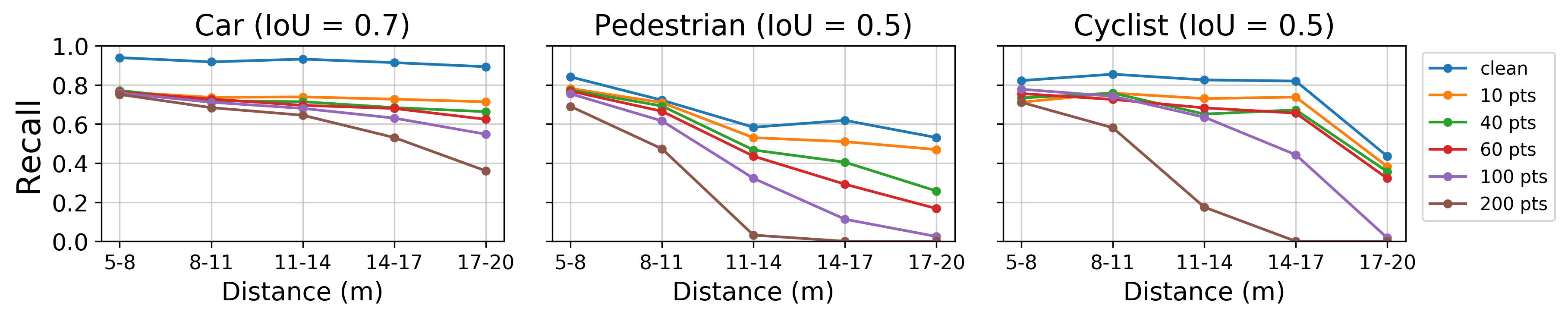}
\includegraphics[width=0.85\textwidth, trim={0 0 0 0}, clip]{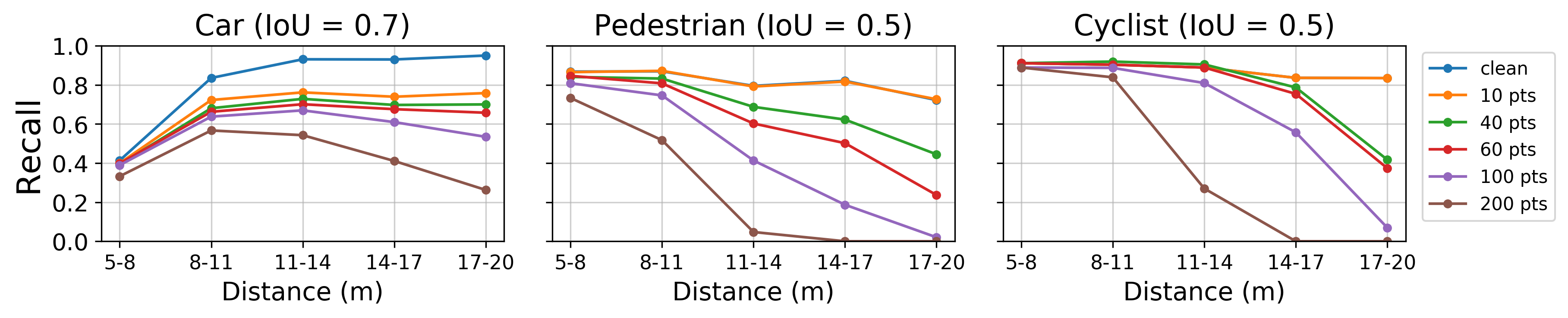}
\vspace{-3.5mm}
\caption{Recall metrics by distance (in metres away from the LiDAR) for different IoU configurations of (top) PointRCNN and (bottom) Point-GNN.}
\label{fig:recall_distance}
\end{figure*}
% \vspace{-3mm}

\section{Experiments \& Results}\label{sec:experiments}
% In this section we describe the dataset and models evaluated, the scenarios considered, and the attacks performed. We then display our results and discuss the findings.

% \subsection{Experimental Setup}
% \subsection{Evaluation Plan}
\vspace{3pt}\vspace{3pt}\noindent\textbf{Models \& Datasets.} The proposed attack was conducted on the validation set (3769 out of 7481 scenes) of the KITTI dataset~\cite{Geiger2013IJRR}. Objects in these scenes are subjected to \textit{ORA-Random} and the resulting perturbed point clouds of the scenes are subsequently passed to popular 3D object detectors to evaluate the performance of the attack. We perform the attacks on three object types,  \textit{Cars}, \textit{Pedestrians} and \textit{Cyclists} as these objects are commonly encountered in AV scenarios. We perform our attack evaluations on widely-used models for 3D point cloud object detection that rely solely on LiDAR data, Point-GNN~\cite{Shi2020PointGNNGN} and PointRCNN~\cite{PRCNN_Shi_2019_CVPR}. The two models differ in how feature extraction is performed for the object detection task. Point-GNN uses a graph neural network that encodes the point-cloud directly as a graph representation for object detection. Whereas, PointRCNN uses a multi-layer perceptron learning approach on point sets to obtain point feature vectors of the point cloud(PointNet++~\cite{qi2017pointnet++}), which are then further processed for object detection.
%We evaluate our proposed attacks on two widely-used models for 3D point cloud object detection that rely solely on LiDAR data. \ignore{hese object detectors exclusively use point clouds as their input.} 

\vspace{-2mm}
\vspace{3pt}\vspace{3pt}\noindent\textbf{Performance Metrics.} For all models and scenarios, we measure the 3D AP and Recall-IOU curves of the models under attack. The 3D AP (average precision) captures the ratio of true positive predictions over all positive predictions and is the primary measure for overall performance of 3D object detectors. The Recall-IOU curve measures the recall of the detector for various IOU thresholds. The goal of the attacker would be to hide real objects from the model, so the measurement of recall is relevant in our case since it captures the ability of the detector to \textbf{not miss} objects. Therefore, $\mathcal{A}$'s goal would be to lower recall scores for target detectors.

%as it is the proportion of real objects that are detected.

\vspace{-2mm}
\vspace{3pt}\vspace{3pt}\noindent\textbf{Evaluation Scenarios.} We consider two scenarios for evaluating the models' performance when under attack. The first is on the performance of the attack applied on the entire KITTI validation dataset. The second evaluation focuses on the impact of ORAs on the detection of \textit{front-near} objects. For both scenarios, we use \textit{ORA-Random} to perturb the point cloud of each individual type of target object found in the scenes with various point-perturbation budgets (10, 20, 40, 60, 100, 150, 200 points) that are within $\mathcal{A}$'s capabilities.
%As we are concerned with the safety implications of the attack, our

\vspace{-2mm}
\subsection{Attack Performance Evaluation}
Table~\ref{table:combined} shows the AP of the 2 models for clean (no point perturbation) and for the various attack budgets used to perturb the 3 object types. The evaluation criteria follows that of the KITTI 3D object detection benchmark (where the detection difficulty levels are determined by the size and occlusion of object). We observed that the AP decreases for increasing $\mathcal{A}_{budget}$. The effect of \textit{ORA-Random} is most significant for \textit{Cyclist} objects and then followed by \textit{Pedestrian} and then \textit{Car} objects. One reason could be that Cyclists are not as common in the dataset compared to the other two classes, resulting in poorer performance. \textit{Pedestrian} and \textit{Cyclist} objects are smaller objects and have significantly fewer points in their point clouds.

% We also studied the Recall-IoU curves for the various point budgets to see the effects of $\mathcal{A}_{budget}$ on recall (i.e. correctly identifying the object) at various IoU thresholds.
From Fig.~\ref{fig:recall_iou}, we observed that when increasing  the point budget, the recall falls. For Cars at IoU $\geq$ 0.7 and for \textit{Pedestrians} and \textit{Cyclists} at IoU $\geq$ 0.5, we observe a significant decline in recall for both clean and attack, with recall falling below 0.5 for most of the attacks, with the exception of the 10-point attack on Point-GNN for pedestrian and cyclist. Our analysis on the KITTI validation set shows that the \textit{ ORA-Random attack is very effective in degrading the object detector's performance and hiding a target object.}

\vspace{-2mm}
\subsection{Attacking Front-Near Objects}
We further investigate whether ORAs can mask front-near objects (objects in close proximity to the ego-vehicle), where accurate detection is critical to the safe operation of the autonomous vehicle. Our results are summarized in Fig.~\ref{fig:recall_distance}. We observe a general trend where objects further away from the LiDAR have lower recall but with the effects being visible even for objects $\leq11m$. The extent of the drop in recall (w.r.t clean) is also correlated to the increase in $\mathcal{A}$'s budget. Noticeably for smaller objects such as \textit{Pedestrian} and \textit{Cyclist}, we observe a higher decrease in recall when increasing $\mathcal{A}_{budget}$ and distance. This is due to the smaller objects' inherent low number of points and its decreasing point density as distance increases. This provides an opportunity for the adversary to use its limited $\mathcal{A}_{budget}$ to perturb a larger proportion of points in the object's point cloud--increasing its success rate of evading detection.

\vspace{-2mm}
\subsection{Discussion \& Implications}
Although \textit{ORA-Random} does not drastically damage the recall for front-near objects, it is still able to significantly lower recall and AP when considering the more general validation dataset. The detection of further away objects remains a critical function especially during the high-speed operation of vehicles. Thus being able to damage model recall in the general case with a random point selection strategy raises grave security concerns. Additionally, the random point selection can be improved upon with more optimized strategies that use methods such as genetic, evolutionary, or Bayesian algorithms \cite{co2019procedural, tu2020physically, zhao2020nudge} to create effective adversarial attacks within a point budget. Overall \textit{ORA-Random} demonstrates that object removal attacks are a real concern which we plan to investigate in depth in future work.
\vspace{-2mm}

\section{Conclusion}% and Future Plans}\label{sec:conclusion}
In this paper, we provide preliminary evidence that with a simple approach of shifting 3D points from a RoI, a LiDAR spoofing adversary is able to effectively perturb the point cloud of a target object to render it undetectable. We performed a sensitivity analysis and found that for smaller objects, the attacks are highly effective at a distance beyond 11m. This poses a safety concern as failure to detect such objects could have life-threatening consequences. In future work, we plan to implement optimization-based point selection ORA strategies, verify the feasibility of ORAs in the physical domain and study the effect of ORA at various distances and driving speeds on AV driving decisions with an AV simulator. As this new class of attack targets a single sensor modality, we are exploring defenses using multi-sensor fusion with RGB cameras.

% The future plans for this work includes the following:
% \vspace{3pt}\noindent$\bullet$ Implement optimization-based point selection strategies to create more effective adversarial attacks within the point budget.
% \vspace{3pt}\noindent$\bullet$ Demonstrate the viability of ORA in the physical domain.

\balance
%\section*{Acknowledgment}



%\bibliographystyle{IEEEtranS}
\bibliographystyle{plain}
\bibliography{biblio}

\end{document}